\documentclass[10pt,twocolumn,letterpaper]{article}

\usepackage[pagenumbers]{wacv} 

\usepackage{float}
\definecolor{wacvblue}{rgb}{0.21,0.49,0.74}
\usepackage[pagebackref,breaklinks,colorlinks,allcolors=wacvblue]{hyperref}

\def\wacvPaperID{3188} 
\def\confName{WACV}
\def\confYear{2026}

\title{GauSSmart: Enhanced 3D Reconstruction through 2D Foundation Models and Geometric Filtering}

\author{Alexander Valverde$^{1*}$\thanks{$^{*}$Project developed while main contributor was a Master student at University of California, Santa Cruz} \quad
Brian Xu$^{2}$ \quad
Yuyin Zhou$^{1}$ \quad
Meng Xu$^{3}$ \quad
Hongyun Wang$^{1}$ \\
$^{1}$University of California, Santa Cruz \quad
$^{2}$Brown University \quad
$^{3}$Kean University \\
}

\begin{document}
\maketitle
\begin{abstract}
Scene reconstruction has emerged as a central challenge in computer vision, with approaches such as Neural Radiance Fields (NeRF) and Gaussian Splatting achieving remarkable progress. While Gaussian Splatting demonstrates strong performance on large-scale datasets, it often struggles to capture fine details or maintain realism in regions with sparse coverage, largely due to the inherent limitations of sparse 3D training data.

In this work, we propose GauSSmart, a hybrid method that effectively bridges 2D foundational models and 3D Gaussian Splatting reconstruction. Our approach integrates established 2D computer vision techniques, including convex filtering and semantic feature supervision from foundational models such as DINO, to enhance Gaussian-based scene reconstruction. By leveraging 2D segmentation priors and high-dimensional feature embeddings, our method guides the densification and refinement of Gaussian splats, improving coverage in underrepresented areas and preserving intricate structural details.

We validate our approach across three datasets, where GauSSmart consistently outperforms existing Gaussian Splatting in the majority of evaluated scenes. Our results demonstrate the significant potential of hybrid 2D-3D approaches, highlighting how the thoughtful combination of 2D foundational models with 3D reconstruction pipelines can overcome the limitations inherent in either approach alone. \textbf{Code is available at:} \url{https://github.com/alevalve/gaussmart}
\end{abstract}
    
\section{Introduction}
\label{sec:intro}

\begin{figure}
    \centering
    \includegraphics[width=\linewidth]{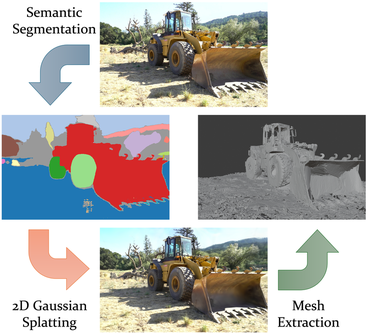}
    \caption{Our 3D reconstruction pipeline: An \textbf{input image} undergoes \textbf{segmentation}, which enhances the \textbf{2D Gaussian splats}, leading to a more geometric \textbf{3D mesh}.}
    \label{fig:teaser}
\end{figure}

Three-dimensional (3D) reconstruction has emerged as a prominent area of research since the introduction of Neural Radiance Fields (NeRF) \cite{mildenhall2020nerf}. More recently, Gaussian Splatting \cite{kerbl2023} has advanced scene reconstruction by exploiting the mathematical properties of 3D Gaussians, which allow for both accurate modeling and efficient rendering. Nevertheless, these methods frequently rely on unstructured point clouds, where uneven density and photogrammetry artifacts often introduce noise, limiting the fidelity of the reconstructed geometry.

In addition, optimization strategies in existing approaches are typically restricted to simple objectives such as L1 reconstruction losses or supervision based on surface normals and depth maps. While these objectives provide a useful training signal, they are insufficient to capture fine-grained, high-frequency details that are critical for realistic scene representation. As a result, reconstructions may lack sharpness and fail to preserve subtle geometric and textural structures present in the original scenes.

Given the spurious conditions of the input point clouds, we first apply a convex outlier removal procedure to eliminate unnecessary points introduced by interpolation errors in COLMAP reconstructions. In addition, we present a framework that leverages 2D foundation models such as DINOv3 \cite{simeoni2025dinov3} and SAM \cite{kirillov2023segment} to guide point cloud densification. Specifically, we increase the density of underrepresented segments by sampling new points according to the covariance structure of the original points within each segment mask. Finally, we introduce a novel loss function built on DINOv3 embeddings, which is differentiable and seamlessly integrates into the Gaussian Splatting pipeline \cite{huang2024gaussian}, enabling improved semantic alignment during optimization.

Our contributions are summarized as follows:

\begin{itemize}
\item \textbf{Convex-guided outlier removal.} We propose a convex-based filtering method that removes spurious points arising from photogrammetry and interpolation errors in COLMAP-generated point clouds.
\item \textbf{Segment-aware point cloud densification.} By leveraging SAM-derived segmentation masks, we regularize and densify sparse regions of the point cloud using a covariance-driven sampling strategy that accounts for both local geometry and segment area.
\item \textbf{Embedding-aligned training.} We introduce a lightweight embedding loss based on DINOv3 features, which enforces consistency between rendered Gaussian appearances and segment-level semantics, thereby improving spatial fidelity and object coherence.
\end{itemize}

\section{Related Works}

\textbf{Outlier Removal.} Outlier removal is a key pre-processing step for noisy point clouds, with two widely used methods being Statistical Outlier Removal (SOR) and Radius Outlier Removal (ROR), both popularized in the Point Cloud Library \cite{Rusu2011PCL,PCL_SOR_ROR}. SOR computes the mean distance of each point to its $k$-nearest neighbors and rejects those deviating from the global mean–standard deviation range, while ROR removes points with fewer than a threshold number of neighbors within a fixed radius. These lightweight filters are usually combined with robust model-fitting methods such as RANSAC \cite{Fischler1981RANSAC} or Trimmed ICP \cite{Chetverikov2002TICP}, surface-fitting and projection approaches like Point Set Surfaces \cite{Alexa2003PSS}, graph-based and variational denoisers \cite{Zeng2019GraphDenoise,Bouaziz2013L0}, and learning-based frameworks such as PointCleanNet \cite{Rakotosaona2019PointCleanNet}.

\textbf{Novel View Synthesis}
Neural Radiance Fields (NeRF)\cite{mildenhall2020nerf} model view-dependent color and density using a multilayer perceptron (MLP) to synthesize novel views. Subsequent variants have improved sampling strategies, scalability, and geometric fidelity\cite{barron2021mipnerf,Liu20neurips_sparse_nerf,Lindell20arxiv_AutoInt,barron2022mipnerf360,yariv2023bakedsdf,cao2024lightning,oechsle2021unisurf,wang2021neus,yariv2021volume,meng_2023_neat,garbin2021fastnerfhighfidelityneuralrendering,lindell2021autoint}. Despite these advances, conventional NeRF models remain computationally expensive and often fail to accurately reconstruct large-scale or texture-poor regions.
More recently, scene representations based on projected Gaussian primitives have enabled fast, high-quality rendering while remaining scalable to large scenes~\cite{kerbl2023,huang2024gaussian,choi2024meshgsadaptivemeshalignedgaussian,Dalal_2024,guédon2023sugarsurfacealignedgaussiansplatting}. In particular, 2D Gaussian Splatting~\cite{huang2024gaussian} improves geometric alignment by orienting primitives along local surface tangents, typically guided by estimated normals. However, coverage gaps may still arise when the initialization is suboptimal or when point density fails to adequately represent fine-scale structures.

\textbf{2D Foundational Models.} Promptable segmentation models such as SAM~\cite{kirillov2023segment} provide strong per-object cues that have been successfully applied in both vision and medical reconstruction~\cite{morshuis2024segmentationguidedmrireconstructionmeaningfully,Ma_2024,li2024anatomaskenhancingmedicalimage,zhou2025distillationlearningguidedimage,shang2024jointsegmentationimagereconstruction}. For 3D Gaussians, SAGA~\cite{cen2025segment3dgaussians} extends SAM-like capabilities but remains limited when handling fine-scale objects. In parallel, semantic backbones such as DINOv2~\cite{oquab2023dinov2} and DINOv3~\cite{simeoni2025dinov3} provide dense, transferable feature embeddings that capture both local detail and global context. In particular, DINOv3 offers semantically consistent representations with strong invariance to viewpoint and appearance changes, making it especially well-suited for embedding-level supervision. This motivates our use of a DINOv3-based loss to complement pixel-domain objectives and enforce semantic alignment in the Gaussian Splatting pipeline.

\textbf{Point-based representations.}
Point clouds are flexible, permutation-invariant 3D carriers for reconstruction~\cite{yu2024pointdreamer,deng2020cvxnet,melaskyriazi2023pc2projectionconditionedpointcloud} and for tasks like 2D projection/detection~\cite{6923303} and point-level segmentation~\cite{qi2017pointnetdeeplearningpoint}. Point-based rendering learns to rasterize point features to images~\cite{rakhimov2022npbgacceleratingneuralpointbased,zhu2024rpbgrobustneuralpointbased}, while self-supervised objectives densify or regularize point sets~\cite{Xiao_2023}. Broader advances span reconstruction~\cite{yu2022pointbertpretraining3dpoint,pang2022maskedautoencoderspointcloud,zhang2022pointm2aemultiscalemaskedautoencoders,dai2021can3dfast3dmedical,xu2023stop3dtargetreconstruction,sun20233dfusionrealtime3dobject,pan2023multidimensionunifiedswintransformer}, completion~\cite{yang2021sanetshuffleattentiondeep,xie2018learningdescriptornetworks3d,zhang2021viewguidedpointcloudcompletion,yu2021pointrdiversepointcloud}, and point operators~\cite{thomas2019kpconvflexibledeformableconvolution,liu2019relationshapeconvolutionalneuralnetwork,Guo_2021}.
\section{Method}

\begin{figure*}[!ht]
    \centering
    \includegraphics[width=\textwidth]{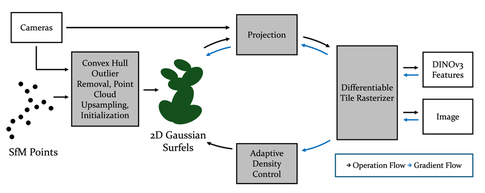}
    \caption{Overview of the GauSSmart pipeline, illustrating the stages of camera clustering, Convex Hull–based outlier removal, point cloud upsampling, and the integration of a DINOv3-based loss.}
    \label{fig:pipeline}
\end{figure*}

\subsection{Convex Outlier Removal}

\begin{figure}[!ht]
    \centering
    \includegraphics[width=\linewidth]{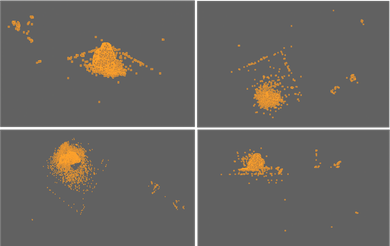}
    \caption{Visualization of a DTU scene point cloud from different viewpoints, showing point cloud outliers even on closed environments}
    \label{fig:outlier}
\end{figure}

Point clouds generated by COLMAP often contain hundreds of spurious points that do not contribute meaningfully to the scene of interest. These outliers typically arise from triangulation errors, introducing noise into the reconstruction. In figure \ref{fig:outlier} we can see external points that do not represent the object of interest. To address this, traditional methods such as Radius Outlier Removal and Statistical Outlier Removal are widely used. These approaches are considered the classical techniques for denoising point cloud representations.

However, most of these methods depend heavily on hyperparameters that are difficult to tune consistently across different scenes, which makes the cleaning process cumbersome. Moreover, they often suffer from limitations: either removing more points than necessary, or failing to eliminate certain outliers due to sparsity issues. Since these techniques generally rely on $k$NN-based strategies that require a minimum number of neighboring points, so they struggle in scenarios with irregular density.

\begin{figure}[!ht]
    \centering
    \includegraphics[width=\linewidth]{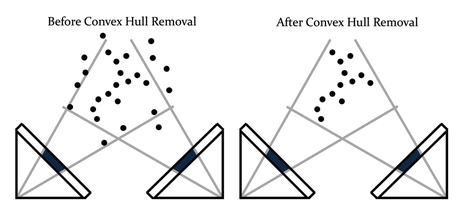}
    \caption{Visualization of the convex hull removal process. Left: the original point cloud containing outliers projected outside the valid viewing cone. Right: the refined point cloud after removing convex hull outliers, which eliminates spurious points and preserves only the consistent geometry.}
    \label{fig:convex_hull}
\end{figure}

To overcome these challenges, we employed a convex hull–based outlier removal method, introduced in \cite{valverde2025convex}, which leverages geometric properties to more reliably filter noise while preserving the integrity of the underlying structure. Unlike previous approaches, this method does not rely on local neighborhoods; instead, it evaluates each point independently and removes it if its distance to the convex hull exceeds a threshold.

\subsection{Camera Clustering and Image Selection}

Large multi-view datasets often contain hundreds of images, making segmentation across all views computationally prohibitive. Drawing inspiration from camera-parameter–driven view-reduction and calibration strategies~\cite{mauro2013overlapping,khan2019fast,yan2024livelearncontinualaction,article,waleed2024cameracalibrationgeometricconstraints}, we propose a clustering-based approach to select a compact, geometry-aware subset of views.

Our method extracts camera centers and forward axes from explicit camera-to-world matrices $\mathrm{c2w}$. To ensure robustness to scene scale and anisotropy, we normalize camera positions by subtracting the mean and dividing by the per-axis standard deviation. We then apply $k$-means clustering to these normalized positions and determine the optimal number of clusters $k$ by evaluating candidate values in the range $k \in [\text{min}, \min(15, \lfloor N/2 \rfloor)]$.

The selection criterion maximizes a composite objective function that balances two key factors: \emph{coverage}, measured as the average of spatial spread and inter-view angular diversity within each cluster, and \emph{compactness}, quantified as negative inertia normalized by the data norm. The objective function is formulated as:

\begin{equation}
\text{score}(k)\;=\;\alpha\,\underbrace{\text{coverage}(k)}_{\text{spread}+\text{angle div.}}
\;+\;\beta\,\underbrace{\text{compactness}(k)}_{-\text{inertia}/\|X\|}.
\end{equation}

After clustering, we select one representative camera per cluster by combining proximity to the de-normalized cluster center with angular uniqueness within the cluster. Each candidate view is scored as the average of two normalized terms: (i) a distance-based component $1/(1+\text{dist})$, and (ii) the mean angular separation to other cluster members, linearly scaled to the interval $[0,1]$. The highest-scoring view from each cluster is selected as the representative.

This approach preserves both spatial coverage and viewpoint diversity while substantially reducing the computational burden by limiting the number of images processed during segmentation. We apply SAM2 large model ~\cite{ravi2024sam2segmentimages} to the selected representative views, generating segmentation masks that include binary regions, bounding boxes, and area measurements. These masks are subsequently used to assign stable segment identifiers to 3D points for downstream processing tasks.

\subsection{Segmentation and Multi-View Projection}

After applying the segmentation process to the respective images to obtain per-view masks, each 3D point is assigned to image segments using a standard projection pipeline ~\cite{6923303}: we normalize coordinates to the dataset scale, transform points into the camera frame using the extrinsics (rotation and translation), and project onto the image plane with the intrinsics, followed by perspective division to obtain pixel locations. For some specific scenarios is required to apply small dataset-specific adjustments (coordinate conventions, calibration differences) and verify projections with visibility checks to ensure accurate point-to-pixel associations across views.

\[
\lambda
\begin{bmatrix} u\\ v\\ 1\end{bmatrix}
= \mathbf{K}\,[\mathbf{R}\;\mathbf{t}]
\begin{bmatrix} \mathbf{P} \\ 1\end{bmatrix},
\qquad
(u,v)=\left(\frac{x}{z},\,\frac{y}{z}\right)
\]

To achieve consistent segment identities across views, we construct a global correspondence map. Segments corresponding to the same object are linked when their 3D point assignments exhibit high normalized overlap. This overlap-based matching yields stable global IDs and robust segment coherence on the fused point cloud, mitigating variability in per-view masks and supporting reliable downstream reconstruction. Importantly, since each 3D point has a unique spatial location, we retain its assignment to the first segment in which it appears. While the same point may project into multiple views, these additional projections provide redundant evidence but no new positional information. Consequently, we discard duplicate assignments, as a point cannot physically belong to multiple objects. This ensures that each point maintains a single, unambiguous label, preserving geometric consistency and preventing fusion artifacts.

\subsection{Mask-Area-Guided Point Cloud Enhancement}

Previous research on point cloud upsampling has demonstrated that increasing point density significantly improves reconstruction quality~\cite{yu2018punetpointcloudupsampling, li2019puganpointcloudupsampling, li2021pointcloudupsamplingdisentangled, rong2024repkpu}. However, uniform densification fails to account for varying importance and visibility across different semantic regions within the scene. Our approach leverages segmentation mask areas from representative views to guide targeted point augmentation. Rather than applying uniform thresholds, we utilize mask area information to estimate appropriate point density for each segment, ensuring visually prominent regions receive adequate geometric representation. For each segment $s_i$ with mask area $A_i$, we determine the target number of points based on the square root relationship between area and required sampling density:

\begin{equation}
n_{\text{target}} = \max\left(\lfloor\sqrt{A_i} \cdot \gamma\rfloor, n_{\min}\right),
\label{eq:target-points}
\end{equation}

where $\gamma = 0.1$ is an empirically determined scaling factor and $n_{\min} = 10$ ensures minimum representation for small segments. For segments where the current point count $|P_{s_i}|$ falls below the target, we compute the augmentation requirement:
\begin{equation}
n_{\text{add}} = n_{\text{target}} - |P_{s_i}|,
\label{eq:points-to-add}
\end{equation}

subject to $n_{\text{add}} > 0$ and the requirement that segments contain at least $n$ existing points to provide sufficient geometric context. The augmentation process generates new points through statistical sampling from existing segment geometry. New point positions $\mathbf{p}_{\text{new}}$ are sampled from the existing distribution with Gaussian noise for geometric diversity:
\begin{equation}
\mathbf{p}_{\text{new}} = \mathbf{p}_{\text{base}} + \boldsymbol{\epsilon}, \quad \mathbf{p}_{\text{base}} \sim P_{s_i}, \quad \boldsymbol{\epsilon} \sim \mathcal{N}(\mathbf{0}, \sigma^2\mathbf{I}),
\label{eq:point-generation}
\end{equation}

where $P_{s_i}$ represents existing points in segment $s_i$, and $\sigma$ is adaptively determined based on local point density. Color attributes for augmented points are derived through interpolation from existing segment points, preserving visual consistency across semantic regions.

\subsection{Feature Embedding Loss}

We build upon recent advances in semantic feature--based supervision~\cite{gong2025dinoslamdinoinformedrgbdslam} by introducing a loss term grounded in DINOv3 feature embeddings~\cite{simeoni2025dinov3}. Traditional photometric losses such as $\ell_1$ or SSIM capture only low-level pixel correspondences and thus fail to encode high-level semantic cues. To address this limitation, we incorporate a DINO-based supervision signal that enforces semantic consistency between ground-truth and rendered images. Prior works typically apply magnitude-based objectives such as $\ell_1$ distance directly on feature vectors, which conflates intensity differences with semantic correspondence. In contrast, we propose a cosine similarity formulation that emphasizes the angular relationship between embeddings, thereby isolating semantic correspondence while remaining invariant to feature magnitude scaling. Formally, given embeddings $f_{\text{gt}}, f_r \in \mathbb{R}^d$ from ground-truth image $I_{\text{gt}}$ and rendered image $I_r$, we define
\begin{equation}
\cos(f_{\text{gt}}, f_r) = \frac{ f_{\text{gt}} \cdot f_r }{ \|f_{\text{gt}}\|_2 \, \|f_r\|_2 },
\end{equation}
with the DINO loss given by
\begin{equation}
L_{\text{DINO}}(f_{\text{gt}}, f_r) = \lambda_{\text{dino}} \cdot \cos(f_{\text{gt}}, f_r),
\end{equation}
where $\lambda_{\text{dino}}$ balances the contribution of semantic supervision. For structural fidelity, we retain a photometric objective composed of an $\ell_1$ term and a structural dissimilarity penalty (DSSIM):
\begin{equation}
L_{\text{photo}} = (1 - \lambda_{\text{dssim}}) \cdot L_{1} + \lambda_{\text{dssim}} \cdot \big(1 - \text{SSIM}(I_{\text{gt}}, I_r)\big).
\end{equation}
The overall training objective combines the semantic and photometric components:
\begin{equation}
L_{\text{total}} = L_{\text{photo}} + L_{\text{DINO}}.
\end{equation}

\begin{figure}[t]
\centering
\setlength{\tabcolsep}{2pt}
\renewcommand{\arraystretch}{1.0}
\begin{tabular}{cccc}
\multicolumn{4}{c}{\textbf{DTU}}\\
\includegraphics[width=0.23\linewidth]{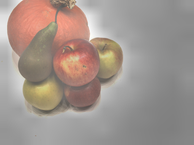} &
\includegraphics[width=0.23\linewidth]{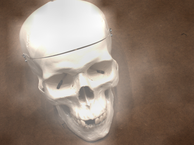} &
\includegraphics[width=0.23\linewidth]{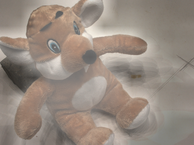} &
\includegraphics[width=0.23\linewidth]{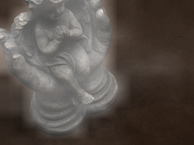} \\[4pt]

\multicolumn{4}{c}{\textbf{Mip-NeRF 360}}\\
\includegraphics[width=0.23\linewidth]{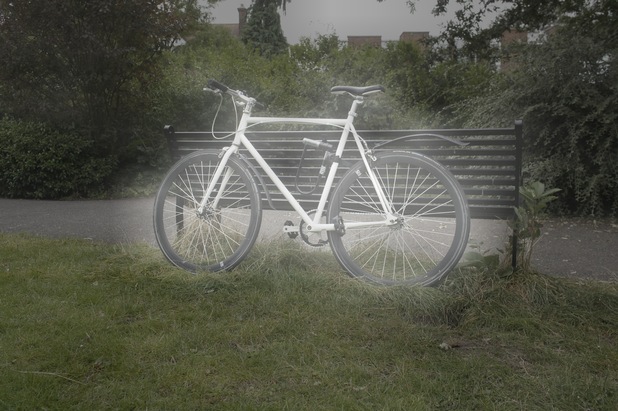} &
\includegraphics[width=0.23\linewidth]{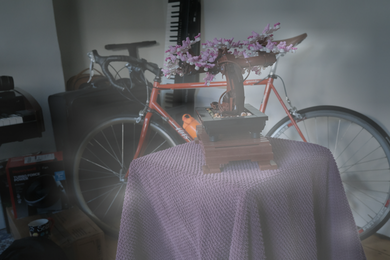} &
\includegraphics[width=0.23\linewidth]{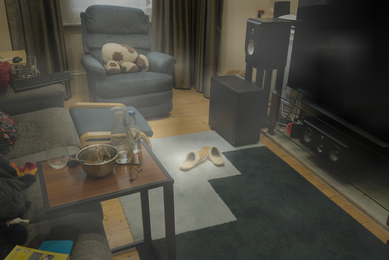} &
\includegraphics[width=0.23\linewidth]{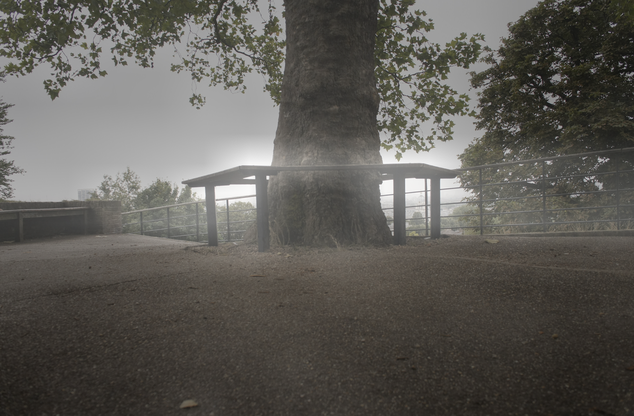} \\[4pt]

\multicolumn{4}{c}{\textbf{T\&T}}\\
\includegraphics[width=0.23\linewidth]{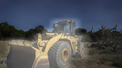} &
\includegraphics[width=0.23\linewidth]{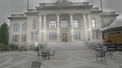} &
\includegraphics[width=0.23\linewidth]{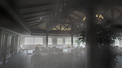} &
\includegraphics[width=0.23\linewidth]{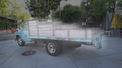} \\
\end{tabular}
\caption{DINOv3 similarity heatmaps across DTU, NeRF, and TYT scenes, highlighting the regions most attended by the model. Brighter areas correspond to stronger alignment with the global embedding.}
\label{fig:dinov3_heatmaps}
\end{figure}

This formulation integrates seamlessly into the 2D Gaussian Splatting framework~\cite{huang2024gaussian}. The additional DINO term complements standard perceptual losses by aligning rendered views with the structural and semantic characteristics of the reference images. Consequently, each Gaussian primitive is optimized not only for photometric accuracy but also for semantic coherence, yielding reconstructions that better reflect the underlying scene semantics.

\section{Experiments}

We now present experiments for our GauSSmart approach, evaluating both appearance and geometry reconstruction across the three benchmark datasets used in this study. Specifically, we conduct experiments on DTU~\cite{jensen2014large}, Mip-NeRF 360~\cite{barron2022mipnerf360}, and Tanks and Temples~\cite{Knapitsch2017}. Our dataset preparation follows the same strategy introduced in the 2DGS paper~\cite{huang2024gaussian}.

\subsection{Implementation Details}
\textbf{Hardware and Environment.} All experiments are conducted on a single NVIDIA RTX A5000 GPU with 24~GB of memory. Our method is implemented in PyTorch~1.12 with CUDA~11.6. Due to hardware differences compared to the original 2DGS implementation (RTX~3090), direct efficiency comparisons are not feasible.

\smallskip
\textbf{Loss Implementation.} We employ DINOv3 with a Vision Transformer base architecture and $16\times16$ patch size (ViT-B/16) as our feature extractor, selected after systematic evaluation across different DINOv3 variants on the Tanks and Temples dataset. The DINO loss weight is set to $\lambda = 0.05$, determined through empirical evaluation over the range $[0.01, 0.9]$ and guided by the weighting strategies described in~\cite{huang2024gaussian}.

\smallskip
\textbf{Gaussian Point Processing.} Our segmentation-aware sampling strategy enforces a minimum threshold of five points per segment during new sample generation. This constraint is critical, as the covariance matrix computation requires sufficient point density to maintain numerical stability and avoid ill-conditioned matrices. Segments falling below this threshold are either merged with neighboring segments or discarded during preprocessing.

\smallskip
\textbf{Outlier Removal.} We adopt the convex-hull–based outlier removal procedure described in~\cite{valverde2025convex}, applying it during both initialization and refinement to maintain geometric consistency across viewpoints.

\smallskip
\textbf{Training Configuration.} Unless otherwise specified, we train for 30{,}000 iterations, following the standard 2DGS optimization schedule.

\section{Results}

Our Gaussmart implementation demonstrates competitive performance across benchmark datasets. Table~\ref{tab:tnt_results} reports a quantitative comparison against 3D Gaussian Splatting (3DGS)\cite{kerbl2023} and 2D Gaussian Splatting (2DGS)\cite{huang2024gaussian}. In terms of PSNR, our method outperforms 2DGS in four of six scenes, indicating reduced noise and improved fidelity. While 2DGS falls short of 3DGS on the Tanks and Temples dataset, the incorporation of our proposed loss and point cloud upsampling narrows this gap.

\begin{table}[t]
\caption{PSNR$\uparrow$, SSIM$\uparrow$, LPIPS$\downarrow$ scores for the Tanks and Temples dataset.}
\label{tab:tnt_results}
\centering
\large
\resizebox{\columnwidth}{!}{%
\begin{tabular}{lcccccc|c}
\toprule
 & Barn & Caterpillar & Courthouse & Ignatius & MeetingRoom & Truck & mean \\
\midrule
\multicolumn{8}{c}{\textbf{PSNR}} \\
\midrule
Gaussmart & 27.20 & 22.83 & 21.61 & 21.60 & 24.90 & 24.19 & 23.72 \\
2DGS      & 27.15 & 22.78 & 21.55 & 21.68 & 24.90 & 24.16 & 23.70 \\
3DGS      & 27.51 & 23.38 & 22.22 & 21.53 & 25.19 & 24.25 & 24.01 \\
\midrule
\multicolumn{8}{c}{\textbf{SSIM}} \\
\midrule
Gaussmart & 0.843 & 0.772 & 0.765 & 0.766 & 0.859 & 0.848 & 0.809 \\
2DGS      & 0.843 & 0.773 & 0.766 & 0.767 & 0.859 & 0.848 & 0.810 \\
3DGS      & 0.852 & 0.791 & 0.779 & 0.776 & 0.866 & 0.853 & 0.820 \\
\midrule
\multicolumn{8}{c}{\textbf{LPIPS}} \\
\midrule
Gaussmart & 0.204 & 0.245 & 0.282 & 0.203 & 0.209 & 0.177 & 0.220 \\
2DGS      & 0.204 & 0.243 & 0.282 & 0.204 & 0.208 & 0.177 & 0.220 \\
3DGS      & 0.160 & 0.190 & 0.266 & 0.153 & 0.141 & 0.108 & 0.170 \\
\bottomrule
\end{tabular}}
\end{table}

Our method surpasses existing approaches on the DTU dataset in PSNR, demonstrating clear improvements in reconstruction quality. The controlled acquisition conditions—high-quality cameras, uniform lighting, and consistent viewpoints—enable our model to fully exploit the available visual information. Sparse initial point clouds in DTU also benefit from our densification strategy, which introduces additional geometric primitives to capture fine details.

\begin{table}[t]
\caption{PSNR$\uparrow$ comparison on DTU dataset.}
\label{tab:dtu_comparison}
\begin{tabular}{lc}
\toprule
Method & PSNR$\uparrow$ \\
\midrule
3DGS & 35.76 \\
SuGaR & 34.57 \\
2DGS & 34.52 \\
Gaussmart (Ours) & \textbf{36.30} \\
\bottomrule
\end{tabular}
\end{table}

These stable imaging conditions provide an ideal setting for our DINO-based loss, allowing semantic features to remain consistent across views. The synergy between controlled capture, point cloud densification, and semantic feature preservation explains the consistent PSNR gains observed on the DTU evaluation set.

\begin{table*}[t]
\centering
\caption{PSNR$\uparrow$, SSIM$\uparrow$, LPIPS$\downarrow$ scores for DTU dataset.}
\label{tab:dtu_results}
\resizebox{\textwidth}{!}{%
\begin{tabular}{lccccccccccccccc|c}
\toprule
 & scan24 & scan37 & scan40 & scan55 & scan63 & scan65 & scan69 & scan83 & scan97 & scan105 & scan106 & scan110 & scan114 & scan118 & scan122 & mean \\
\midrule
\multicolumn{17}{c}{\textbf{PSNR}} \\
\midrule
Gaussmart & 36.42 & 32.99 & 35.57 & 35.54 & 38.98 & 36.15 & 34.69 & 35.86 & 33.68 & 36.72 & 37.63 & 35.96 & 34.65 & 39.22 & 39.07 & 36.30 \\
\midrule
\multicolumn{17}{c}{\textbf{SSIM}} \\
\midrule
Gaussmart & 0.964 & 0.962 & 0.962 & 0.938 & 0.975 & 0.936 & 0.940 & 0.921 & 0.929 & 0.933 & 0.941 & 0.934 & 0.936 & 0.941 & 0.937 & 0.941 \\
\midrule
\multicolumn{17}{c}{\textbf{LPIPS}} \\
\midrule
Gaussmart & 0.062 & 0.091 & 0.093 & 0.146 & 0.075 & 0.157 & 0.132 & 0.223 & 0.177 & 0.174 & 0.209 & 0.225 & 0.192 & 0.225 & 0.239 & 0.165 \\
\bottomrule
\end{tabular}}
\end{table*}

\begin{table}[H]
\centering
\caption{PSNR$\uparrow$, SSIM$\uparrow$, LPIPS$\downarrow$ scores for the MipNeRF360 dataset.}
\large
\resizebox{\columnwidth}{!}{%
\begin{tabular}{lccccccccc|c}
\toprule
 & bicycle & flowers & garden & stump & treehill & room & counter & kitchen & bonsai & mean \\
\midrule
\multicolumn{11}{c}{\textbf{PSNR}} \\
\midrule
Gaussmart & 24.78 & 21.05 & 26.96 & 26.41 & 22.41 & 31.51 & 28.82 & 31.18 & 31.94 & 27.23 \\
2DGS      & 24.87 & 21.15 & 26.95 & 26.47 & 22.27 & 31.06 & 28.55 & 30.50 & 31.52 & 26.93 \\
3DGS      & 25.24 & 21.52 & 27.41 & 25.07 & 22.49 & 30.63 & 28.70 & 30.32 & 31.98 & 27.26 \\
\midrule
\multicolumn{11}{c}{\textbf{SSIM}} \\
\midrule
Gaussmart & 0.725 & 0.570 & 0.841 & 0.760 & 0.622 & 0.921 & 0.960 & 0.925 & 0.940 & 0.807 \\
2DGS      & 0.752 & 0.588 & 0.852 & 0.765 & 0.627 & 0.912 & 0.900 & 0.919 & 0.933 & 0.805 \\
3DGS      & 0.771 & 0.605 & 0.868 & 0.775 & 0.638 & 0.914 & 0.905 & 0.922 & 0.938 & 0.815 \\
\midrule
\multicolumn{11}{c}{\textbf{LPIPS}} \\
\midrule
Gaussmart & 0.282 & 0.382 & 0.150 & 0.266 & 0.380 & 0.206 & 0.198 & 0.129 & 0.188 & 0.242 \\
2DGS      & 0.218 & 0.346 & 0.115 & 0.222 & 0.329 & 0.223 & 0.208 & 0.133 & 0.214 & 0.223 \\
3DGS      & 0.205 & 0.336 & 0.103 & 0.210 & 0.317 & 0.220 & 0.204 & 0.129 & 0.205 & 0.214 \\
\bottomrule
\end{tabular}}
\end{table}

\begin{figure*}[t]\centering
    \begin{tabular}{cccc}
        GT                                                 & Ours                                               & 2DGS                                    & 3DGS                                                          \\ 
        \includegraphics[width=0.20\linewidth]{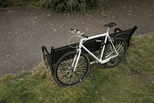}   & \includegraphics[width=0.20\linewidth]{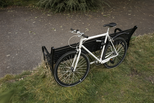}   & \includegraphics[width=0.20\linewidth]{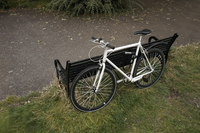}   & \includegraphics[width=0.20\linewidth]{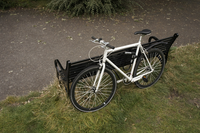}  \\
        \includegraphics[width=0.20\linewidth]{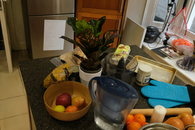}   & \includegraphics[width=0.20\linewidth]{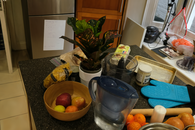}   & \includegraphics[width=0.20\linewidth]{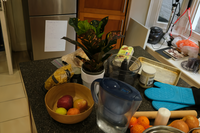}   & \includegraphics[width=0.20\linewidth]{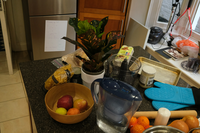}  \\
        \includegraphics[width=0.20\linewidth]{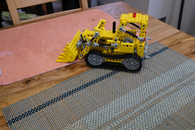}   & \includegraphics[width=0.20\linewidth]{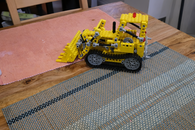}   & \includegraphics[width=0.20\linewidth]{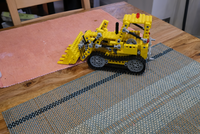}   & \includegraphics[width=0.20\linewidth]{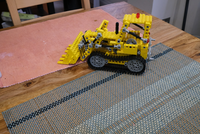}  \\
        \includegraphics[width=0.20\linewidth]{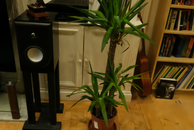}   & \includegraphics[width=0.20\linewidth]{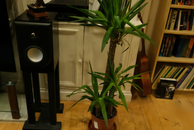}   & \includegraphics[width=0.20\linewidth]{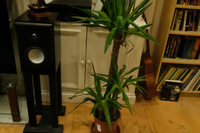}   & \includegraphics[width=0.20\linewidth]{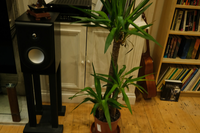}  \\
        \includegraphics[width=0.20\linewidth]{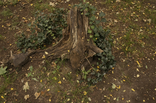}   & \includegraphics[width=0.20\linewidth]{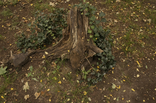}   & \includegraphics[width=0.20\linewidth]{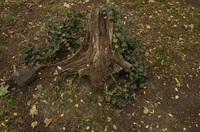}   & \includegraphics[width=0.20\linewidth]{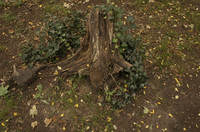}  \\
        \includegraphics[width=0.20\linewidth]{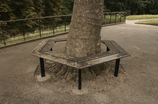}   & \includegraphics[width=0.20\linewidth]{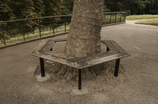}   & \includegraphics[width=0.20\linewidth]{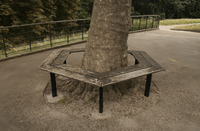}   & \includegraphics[width=0.20\linewidth]{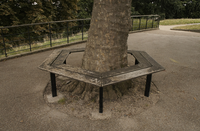}  \\
    \end{tabular}
    \caption{A side-by-side comparison of results from our method and the baselines.}
    \label{fig:results_nerf}
\end{figure*}

In Fig.~\ref{fig:results_nerf}, our results are comparable to those of 3DGS, while exhibiting noticeably less noise than 2DGS. This improvement stems from reducing noise in the underlying point cloud, which often introduces bubble-like artifacts and leads to noisy renderings. Our reconstructions also preserve colors more consistently, confirming that the DINO-based loss captures and retains high-level semantic features. For outdoor scenes, our approach outperforms 2DGS across multiple cases, demonstrating strong generalization beyond indoor environments. This is particularly evident in the \textit{treehill} scene, where complex structures such as root systems and tree trunks benefit from our feature-aware loss.

For indoor environments, \textbf{GauSSmart} consistently achieves superior performance across all three quantitative metrics, surpassing both 2DGS and 3DGS. This improvement highlights the robustness of our method under controlled lighting conditions, where detailed textures and diverse scene elements are more effectively reconstructed.

\begin{figure*}[t]\centering
    \begin{tabular}{cccc}
        GT                                                 & Ours                                               & 2DGS                                    & 3DGS                                                          \\ 
        \includegraphics[width=0.20\linewidth]{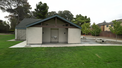}   & \includegraphics[width=0.20\linewidth]{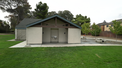}   & \includegraphics[width=0.20\linewidth]{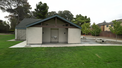}   & \includegraphics[width=0.20\linewidth]{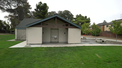}  \\
        \includegraphics[width=0.20\linewidth]{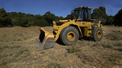}   & \includegraphics[width=0.20\linewidth]{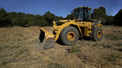}   & \includegraphics[width=0.20\linewidth]{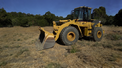}   & \includegraphics[width=0.20\linewidth]{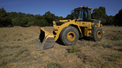}  \\
        \includegraphics[width=0.20\linewidth]{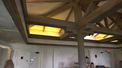}   & \includegraphics[width=0.20\linewidth]{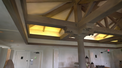}   & \includegraphics[width=0.20\linewidth]{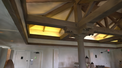}   & \includegraphics[width=0.20\linewidth]{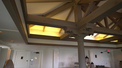}  \\
        \includegraphics[width=0.20\linewidth]{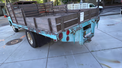}   & \includegraphics[width=0.20\linewidth]{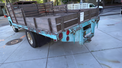}   & \includegraphics[width=0.20\linewidth]{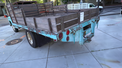}   & \includegraphics[width=0.20\linewidth]{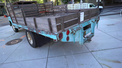}  \\
        
    \end{tabular}
    \caption{A side-by-side comparison of results from our method and the baselines.}
    \label{fig:results_tnt}
\end{figure*}

Qualitative comparisons in Fig.~\ref{fig:results_tnt} further confirm the benefits of our approach. Structural elements such as the walls and sky regions in the Caterpillar scene are reconstructed more cleanly and stably than with 2DGS or 3DGS. While 2DGS introduces noisy textures and bubble-like artifacts in low-texture areas, Gaussmart suppresses such instabilities, producing smoother surfaces and more natural transitions. Beyond large homogeneous regions, our method also demonstrates sharper preservation of edges and fine object boundaries, as in the courthouse scene.

To further assess generalization, we conduct experiments on the DTU dataset, which consists of object-centric scenes captured under controlled conditions. As shown in Table~\ref{tab:dtu_comparison}, our method outperforms recent baselines, including 2DGS, 3DGS, and SuGaR, achieving a PSNR improvement of nearly one point over the best competing approach. This highlights the ability of our model to produce cleaner and more faithful reconstructions. We attribute these improvements to the synergy between DTU’s multi-view acquisition and our DINO-based loss, which leverages high-level features to enhance geometry-aware consistency.

Per-scene results in Table~\ref{tab:dtu_results} further support this conclusion. Gaussmart achieves a mean PSNR of 36.30 dB, with several scans (e.g., 63, 118, 122) exceeding 39 dB, demonstrating robustness across diverse geometries.
\section{Ablations}

Table~\ref{tab:ablations} presents the mean PSNR across ablation configurations on Tanks and Temples.

\begin{table}[t]
\caption{Ablation study results applied on Tanks and Temples Training Scenes (mean PSNR).}
\label{tab:ablations}
\begin{tabular}{lc}
\toprule
\textbf{Experiment} & \textbf{Mean PSNR} \\
\midrule
DINO/7000 loss      & 23.634 \\
DINO/3000 loss      & 23.611 \\
Just PC Improvement & 23.611 \\
Hull/PC Improvement & 23.637 \\
Full model          & \textbf{23.641} \\
\bottomrule
\end{tabular}
\end{table}

Point cloud improvements alone achieve limited gains, indicating that densification in isolation is insufficient. Combining hull filtering with point cloud improvements increases mean PSNR, showing that geometric outlier removal stabilizes 3D reconstructions. Our complete model, which integrates DINO guidance with hull filtering and point cloud improvements, attains optimal performance, confirming the complementary benefits of semantic loss and geometric refinement.

\subsection{Limitations}

Despite the promising results, our approach presents certain computational challenges that warrant consideration. The necessity to compute DINO embeddings each time an image passes through the pipeline introduces a significant computational overhead, potentially increasing processing time by up to 1.5x depending on the size and complexity of the scene. Additionally, our approach computes SAM maps for each image during processing, which adds further computational burden. 

Through our experiments, we discovered consistent patterns across datasets: DTU scenes consistently utilize 3 images given that the cameras are the same for all objects, while Mip-NeRF 360 exhibits a similar pattern with 15 images. This observation suggests a promising avenue for optimization—storing that could significantly reduce training time, that is the main drawback of the current approach.

\section{Conclusions}

In this work, we introduced GauSSmart, a novel approach that combines the strengths of 2D foundational models and 3D methods. Our method highlights the potential of integrating established 2D computer vision techniques, including convex filtering and other foundational components, with 3D reconstruction pipelines. Experimental evaluation across three diverse datasets shows that GauSSmart consistently outperforms existing Gaussian Splatting methods in most scenes, supporting our hypothesis that hybrid 2D–3D approaches can effectively leverage the complementary strengths of both domains. Its superior performance stems from the strategic use of robust 2D priors to compensate for the limitations of sparse 3D training data, underscoring a key insight for the field: the scarcity of high-quality 3D data necessitates intelligent integration with rich 2D image representations.

Looking ahead, the recent release of DiNOv3 presents promising oportunities for advancing 2D foundational models in 3D reconstruction tasks. The enhanced capabilities and improved feature representations offered by DiNOv3 could unlock further greater performance gains when integrated into Gaussian Splatting frameworks.

\onecolumn

\section*{Supplementary Materials}

\subsection*{Geometric Results}

\captionof{table}{DTU evaluation results using 2DGS pipeline. Metrics are mean distance-to-surface (d2s), mean surface-to-distance (s2d), and overall error.}
\begin{tabular}{lccc}
\toprule
Scan & mean\_d2s & mean\_s2d & overall \\
\midrule
24  & 2.6143 & 1.4773 & 2.0458 \\
37  & 1.8802 & 0.8228 & 1.3515 \\
40  & 2.0464 & 1.6955 & 1.8709 \\
55  & 1.6924 & 0.6944 & 1.1934 \\
63  & 2.8613 & 2.6491 & 2.7552 \\
65  & 2.7081 & 1.8333 & 2.2707 \\
69  & 2.0261 & 1.1439 & 1.5850 \\
83  & 2.2375 & 1.8825 & 2.0600 \\
97  & 1.9723 & 1.8632 & 1.9178 \\
105 & 1.9679 & 1.4783 & 1.7231 \\
106 & 2.4727 & 1.1019 & 1.7873 \\
110 & 2.6417 & 1.2023 & 1.9220 \\
114 & 1.5179 & 0.9364 & 1.2272 \\
118 & 1.8501 & 0.7577 & 1.3039 \\
122 & 2.1410 & 0.8514 & 1.4962 \\
\bottomrule
\end{tabular}

\begin{table}[h!]
\caption{Indoor scenes (DTU, Nerf indoor, Meetingroom).}
\begin{tabular}{lcc}
\toprule
Category & Mean Points Added & Mean Segments with New Points \\
\midrule
Indoor Scenes & 116.08 & 8.44 \\
\bottomrule
\end{tabular}
\end{table}

\begin{table}[h!]
\caption{Outdoor scenes (Tanks except Meetingroom, Nerf outdoor).}
\begin{tabular}{lcc}
\toprule
Category & Mean Points Added & Mean Segments with New Points \\
\midrule
Outdoor Scenes & 88.15 & 9.36 \\
\bottomrule
\end{tabular}
\end{table}

\subsection*{Camera Clustering}

Examples of the images selected by clustering method for a set of scenes from DTU, MipNerf360 and Tanks and Temples

\begin{figure*}[t]\centering
    \begin{tabular}{cccc}
        000045.jpg & 000050.jpg & 000088.jpg & 000147.jpg \\
        \includegraphics[width=0.20\linewidth]{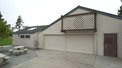} & 
        \includegraphics[width=0.20\linewidth]{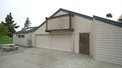} & 
        \includegraphics[width=0.20\linewidth]{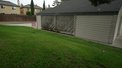} & 
        \includegraphics[width=0.20\linewidth]{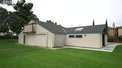} \\
        
        000003.jpg & 000009.jpg & 000037.jpg & 000111.jpg \\
        \includegraphics[width=0.20\linewidth]{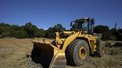} & 
        \includegraphics[width=0.20\linewidth]{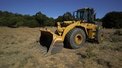} & 
        \includegraphics[width=0.20\linewidth]{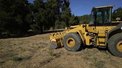} & 
        \includegraphics[width=0.20\linewidth]{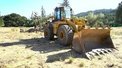} \\
        
        000112.jpg & & & \\
        \includegraphics[width=0.20\linewidth]{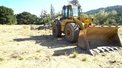} & & & \\
    \end{tabular}
    \caption{Selected clustered camera views from Barn and Caterpillar scenes.}
    \label{fig:results_camera_clustering}
\end{figure*}

\begin{figure*}[t]\centering
    \begin{tabular}{ccc}
        0000.png & 0004.png & 0045.png \\
        \includegraphics[width=0.25\linewidth]{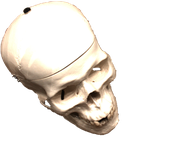} & 
        \includegraphics[width=0.25\linewidth]{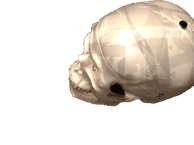} & 
        \includegraphics[width=0.25\linewidth]{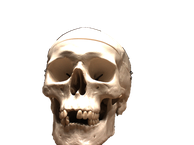} \\
        
        0000.png & 0011.png & 0019.png \\
        \includegraphics[width=0.25\linewidth]{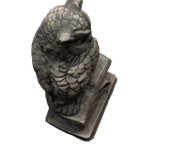} & 
        \includegraphics[width=0.25\linewidth]{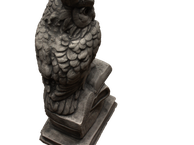} & 
        \includegraphics[width=0.25\linewidth]{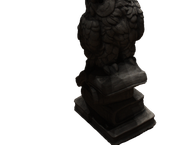} \\
    \end{tabular}
    \caption{Clustered camera views from DTU scans: Scan65 (top row) and Scan122 (bottom row).}
    \label{fig:results_dtu_clustering}
\end{figure*}

\begin{figure*}[t]\centering
\setlength{\tabcolsep}{2pt}
\renewcommand{\arraystretch}{0}
\begin{tabular}{ccccc}
\multicolumn{5}{c}{\textbf{Bicycle}}\\[2pt]
\includegraphics[width=0.18\linewidth]{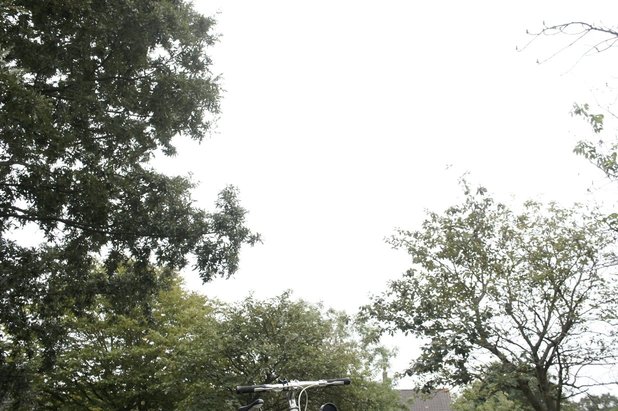} &
\includegraphics[width=0.18\linewidth]{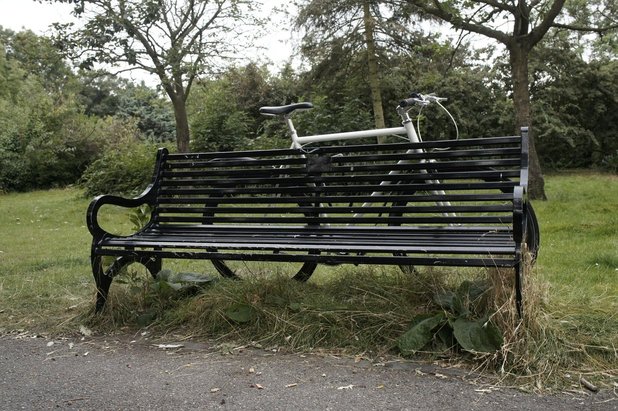} &
\includegraphics[width=0.18\linewidth]{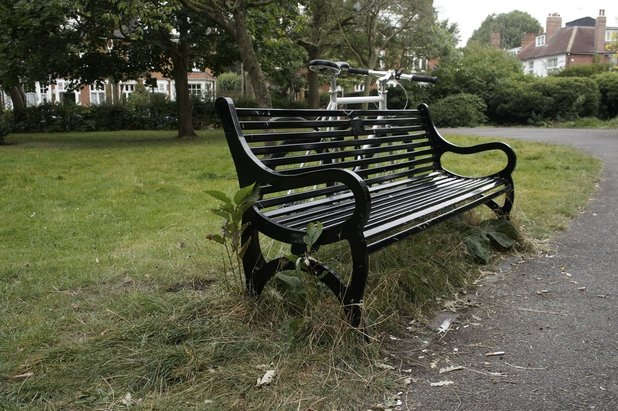} &
\includegraphics[width=0.18\linewidth]{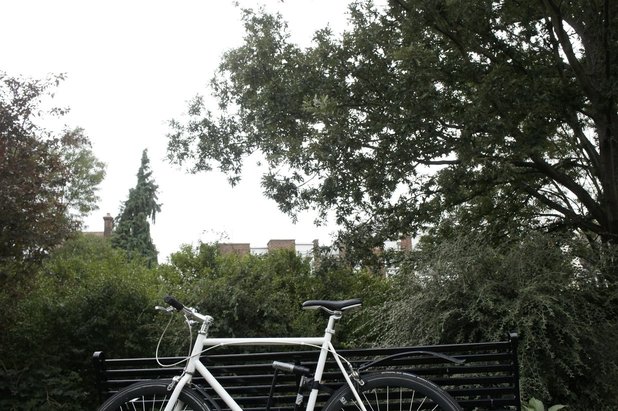} &
\includegraphics[width=0.18\linewidth]{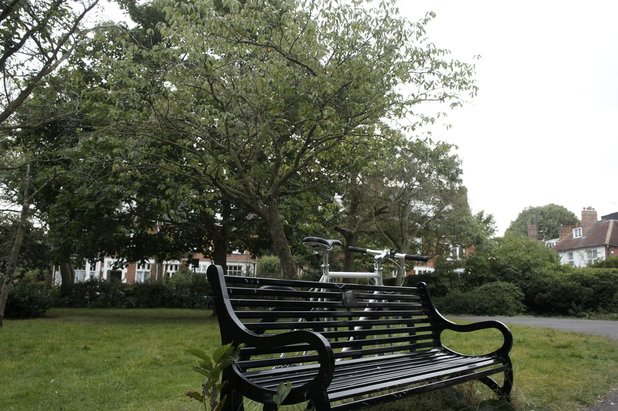} \\
\includegraphics[width=0.18\linewidth]{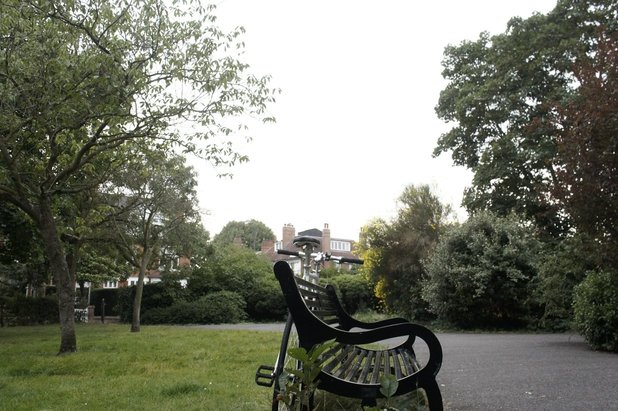} &
\includegraphics[width=0.18\linewidth]{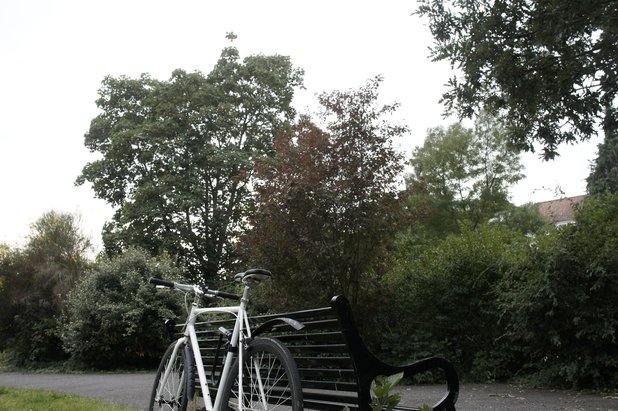} &
\includegraphics[width=0.18\linewidth]{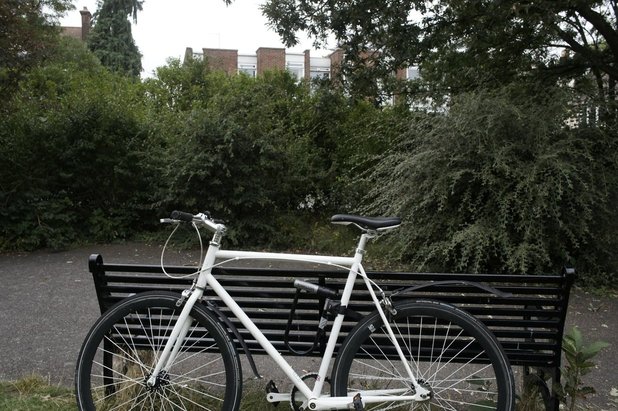} &
\includegraphics[width=0.18\linewidth]{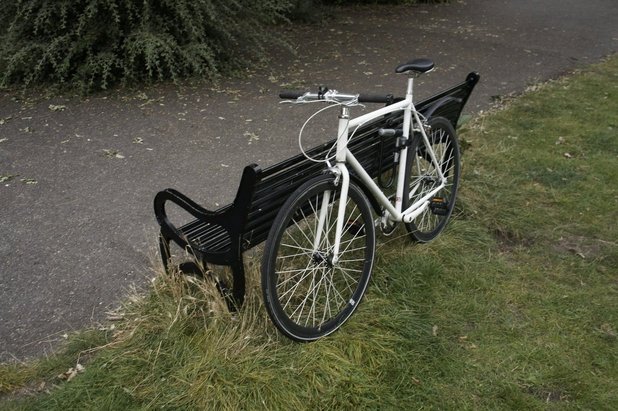} &
\includegraphics[width=0.18\linewidth]{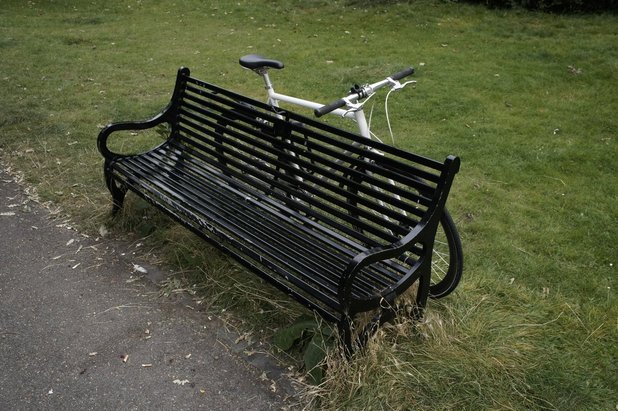} \\
\includegraphics[width=0.18\linewidth]{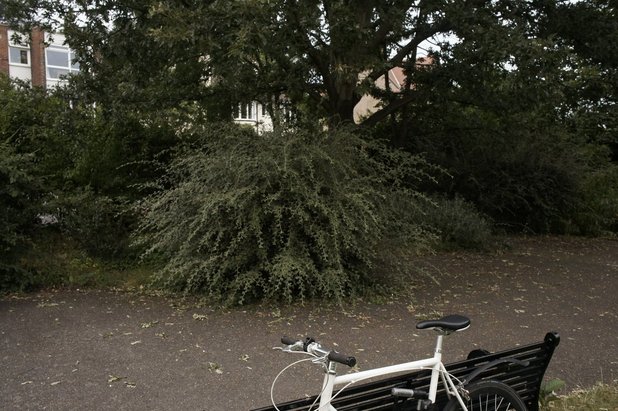} &
\includegraphics[width=0.18\linewidth]{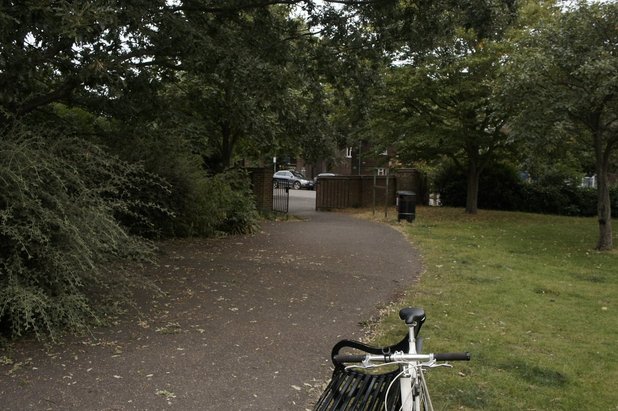} &
\includegraphics[width=0.18\linewidth]{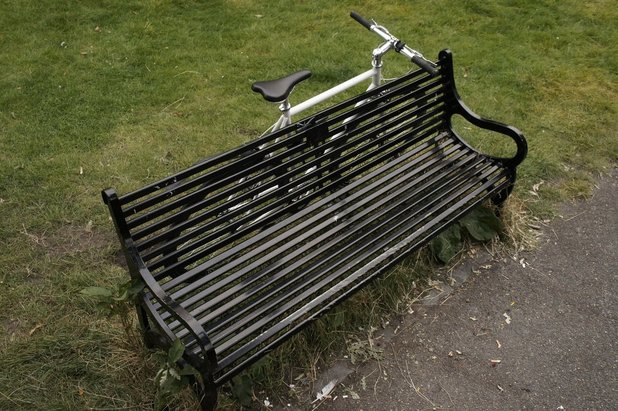} &
\includegraphics[width=0.18\linewidth]{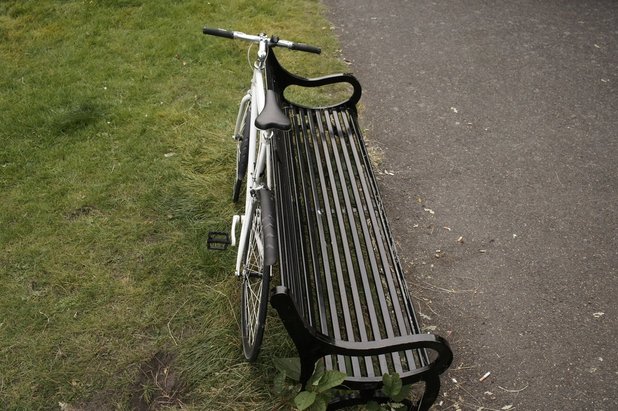} &
\includegraphics[width=0.18\linewidth]{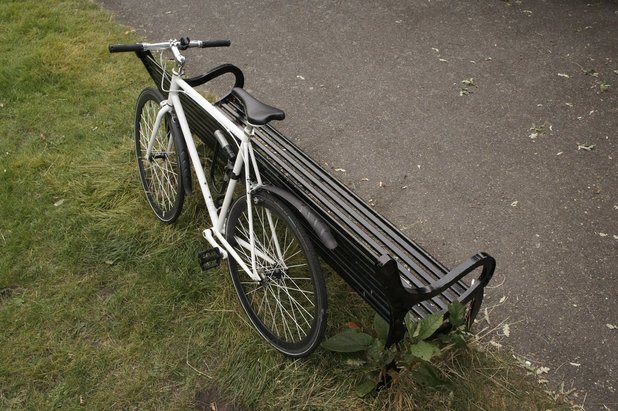} \\
\\[3pt]
\multicolumn{5}{c}{\textbf{Counter}}\\[2pt]
\includegraphics[width=0.18\linewidth]{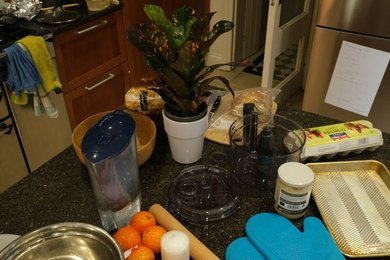} &
\includegraphics[width=0.18\linewidth]{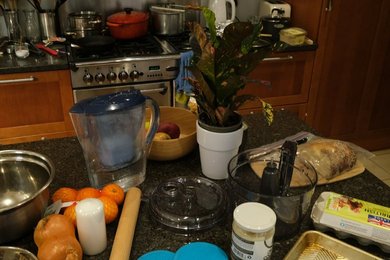} &
\includegraphics[width=0.18\linewidth]{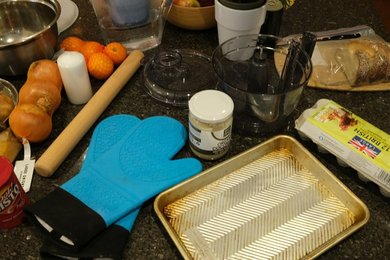} &
\includegraphics[width=0.18\linewidth]{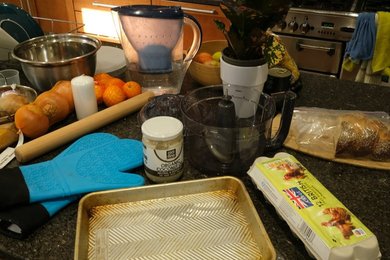} &
\includegraphics[width=0.18\linewidth]{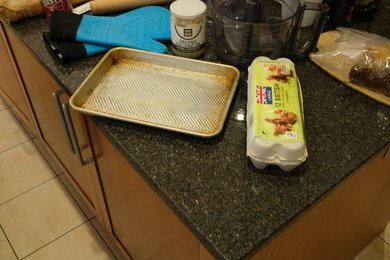} \\
\includegraphics[width=0.18\linewidth]{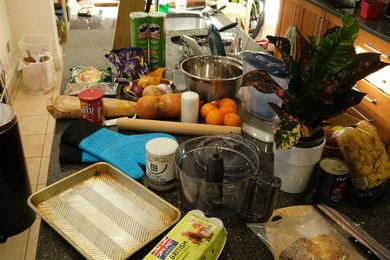} &
\includegraphics[width=0.18\linewidth]{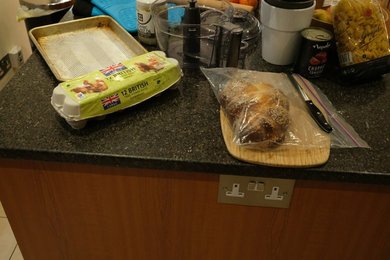} &
\includegraphics[width=0.18\linewidth]{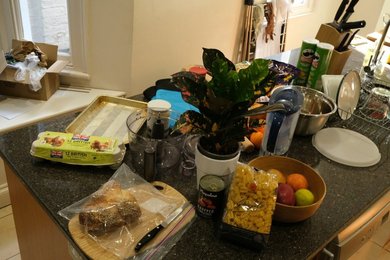} &
\includegraphics[width=0.18\linewidth]{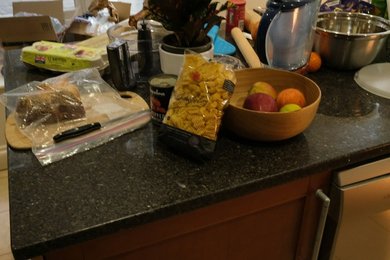} &
\includegraphics[width=0.18\linewidth]{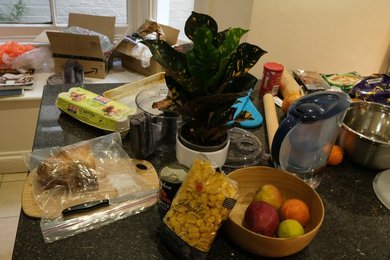} \\
\includegraphics[width=0.18\linewidth]{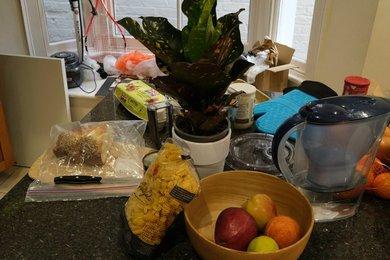} &
\includegraphics[width=0.18\linewidth]{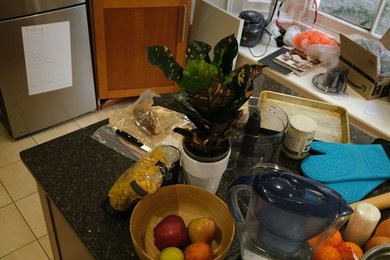} &
\includegraphics[width=0.18\linewidth]{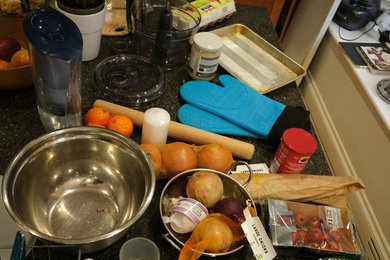} &
\includegraphics[width=0.18\linewidth]{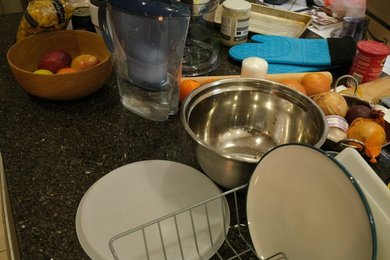} &
\includegraphics[width=0.18\linewidth]{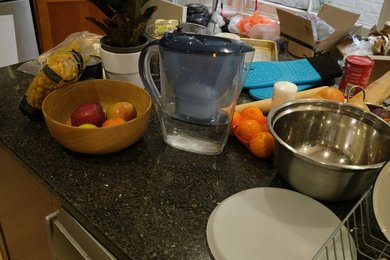} \\
\end{tabular}
\caption{Selected clustered camera views — \textbf{Bicycle} (top) and \textbf{Counter} (bottom).}
\label{fig:camera_clustering_bicycle_counter}
\end{figure*}

{
    \small
    \clearpage
    \bibliographystyle{ieeenat_fullname}
    \bibliography{main}

@String(CVPR= {IEEE Conf. Comput. Vis. Pattern Recog.})

@String(ECCV= {Eur. Conf. Comput. Vis.})

@String(ICPR = {Int. Conf. Pattern Recog.})

@String(BMVC= {Brit. Mach. Vis. Conf.})

@String(TOG= {ACM Trans. Graph.})

@String(ICME = {Int. Conf. Multimedia and Expo})

@String(CVPR  = {CVPR})

@String(ECCV  = {ECCV})

@String(ICPR  = {ICPR})

@String(BMVC  =	{BMVC})

@String(TOG   = {ACM TOG})

@String(ICME  =	{ICME})

@article{huang2024gaussian,
  title={2D Gaussian Splatting for Geometrically Accurate Radiance Fields},
  author={Huang, Binbin and Yu, Zehao and Chen, Anpei and Geiger, Andreas},
  journal={ACM Transactions on Graphics (TOG)},
  volume={43},
  number={4},
  pages={32},
  year={2024},
  publisher={ACM}
}

@article{kerbl2023,
  title={3D Gaussian Splatting for Real-Time Radiance Field Rendering},
  author={Kerbl, Bernhard and Kopanas, Georgios and Leimk{\"u}hler, Thomas and Drettakis, George},
  journal={ACM Transactions on Graphics (TOG)},
  volume={42},
  number={4},
  pages={1--14},
  year={2023},
  publisher={ACM}
}

@article{deng2020cvxnet,
  title={CvxNet: Learnable Convex Decomposition},
  author={Deng, Boyang and Genova, Kyle and Yazdani, Soroosh and Bouaziz, Sofien and Hinton, Geoffrey and Tagliasacchi, Andrea},
  journal={arXiv preprint arXiv:1909.05736},
  year={2020}
}

@article{yu2024pointdreamer,
  title={{PointDreamer: Zero-shot 3D Textured Mesh Reconstruction from Colored Point Cloud by 2D Inpainting}},
  author={Yu, Qiao and Li, Xianzhi and Tang, Yuan and Xu, Jinfeng and Hu, Long and Hao, Yixue and Chen, Min},
  journal={arXiv preprint arXiv:2406.15811},
  year={2024},
  note={\url{https://arxiv.org/abs/2406.15811}}
}

@inproceedings{mildenhall2020nerf,
  title={NeRF: Representing Scenes as Neural Radiance Fields for View Synthesis},
  author={Ben Mildenhall and Pratul P. Srinivasan and Matthew Tancik and Jonathan T. Barron and Ravi Ramamoorthi and Ren Ng},
  year={2020},
  booktitle={ECCV},
}

@inproceedings{jensen2014large,
  title={Large scale multi-view stereopsis evaluation},
  author={Jensen, Rasmus and Dahl, Anders and Vogiatzis, George and Tola, Engil and Aan{\ae}s, Henrik},
  booktitle={2014 IEEE Conference on Computer Vision and Pattern Recognition},
  pages={406--413},
  year={2014},
  organization={IEEE}
}

@misc{kirillov2023segment,
      title={Segment Anything}, 
      author={Alexander Kirillov and Eric Mintun and Nikhila Ravi and Hanzi Mao and Chloe Rolland and Laura Gustafson and Tete Xiao and Spencer Whitehead and Alexander C. Berg and Wan-Yen Lo and Piotr Dollár and Ross Girshick},
      year={2023},
      eprint={2304.02643},
      archivePrefix={arXiv},
      primaryClass={cs.CV},
      url={https://arxiv.org/abs/2304.02643}, 
}

@misc{choi2024meshgsadaptivemeshalignedgaussian,
      title={MeshGS: Adaptive Mesh-Aligned Gaussian Splatting for High-Quality Rendering}, 
      author={Jaehoon Choi and Yonghan Lee and Hyungtae Lee and Heesung Kwon and Dinesh Manocha},
      year={2024},
      eprint={2410.08941},
      archivePrefix={arXiv},
      primaryClass={cs.CV},
      url={https://arxiv.org/abs/2410.08941}, 
}

@inproceedings{lindell2021autoint,
author = {Lindell, David B. and Martel, Julien N.P. and Wetzstein, Gordon},
title = {AutoInt: Automatic Integration for Fast Neural Volume Rendering},
booktitle = {Proceedings of the conference on Computer Vision and Pattern Recognition (CVPR)},
year={2021}
}

@misc{garbin2021fastnerfhighfidelityneuralrendering,
      title={FastNeRF: High-Fidelity Neural Rendering at 200FPS}, 
      author={Stephan J. Garbin and Marek Kowalski and Matthew Johnson and Jamie Shotton and Julien Valentin},
      year={2021},
      eprint={2103.10380},
      archivePrefix={arXiv},
      primaryClass={cs.CV},
      url={https://arxiv.org/abs/2103.10380}, 
}

@article{Dalal_2024,
   title={Gaussian Splatting: 3D Reconstruction and Novel View Synthesis: A Review},
   volume={12},
   ISSN={2169-3536},
   url={http://dx.doi.org/10.1109/ACCESS.2024.3408318},
   DOI={10.1109/access.2024.3408318},
   journal={IEEE Access},
   publisher={Institute of Electrical and Electronics Engineers (IEEE)},
   author={Dalal, Anurag and Hagen, Daniel and Robbersmyr, Kjell G. and Knausgård, Kristian Muri},
   year={2024},
   pages={96797–96820} }

@misc{melaskyriazi2023pc2projectionconditionedpointcloud,
      title={$PC^2$: Projection-Conditioned Point Cloud Diffusion for Single-Image 3D Reconstruction}, 
      author={Luke Melas-Kyriazi and Christian Rupprecht and Andrea Vedaldi},
      year={2023},
      eprint={2302.10668},
      archivePrefix={arXiv},
      primaryClass={cs.CV},
      url={https://arxiv.org/abs/2302.10668}, 
}

@INPROCEEDINGS{6923303,
  author={Chmelar, Pavel and Beran, Ladislav and Kudriavtseva, Nataliia},
  booktitle={Proceedings ELMAR-2014}, 
  title={Projection of point cloud for basic object detection}, 
  year={2014},
  volume={},
  number={},
  pages={1-4},
  doi={10.1109/ELMAR.2014.6923303}}

@misc{qi2017pointnetdeeplearningpoint,
      title={PointNet: Deep Learning on Point Sets for 3D Classification and Segmentation}, 
      author={Charles R. Qi and Hao Su and Kaichun Mo and Leonidas J. Guibas},
      year={2017},
      eprint={1612.00593},
      archivePrefix={arXiv},
      primaryClass={cs.CV},
      url={https://arxiv.org/abs/1612.00593}, 
}

@misc{guédon2023sugarsurfacealignedgaussiansplatting,
      title={SuGaR: Surface-Aligned Gaussian Splatting for Efficient 3D Mesh Reconstruction and High-Quality Mesh Rendering}, 
      author={Antoine Guédon and Vincent Lepetit},
      year={2023},
      eprint={2311.12775},
      archivePrefix={arXiv},
      primaryClass={cs.GR},
      url={https://arxiv.org/abs/2311.12775}, 
}

@misc{rakhimov2022npbgacceleratingneuralpointbased,
      title={NPBG++: Accelerating Neural Point-Based Graphics}, 
      author={Ruslan Rakhimov and Andrei-Timotei Ardelean and Victor Lempitsky and Evgeny Burnaev},
      year={2022},
      eprint={2203.13318},
      archivePrefix={arXiv},
      primaryClass={cs.CV},
      url={https://arxiv.org/abs/2203.13318}, 
}

@misc{zhu2024rpbgrobustneuralpointbased,
      title={RPBG: Towards Robust Neural Point-based Graphics in the Wild}, 
      author={Qingtian Zhu and Zizhuang Wei and Zhongtian Zheng and Yifan Zhan and Zhuyu Yao and Jiawang Zhang and Kejian Wu and Yinqiang Zheng},
      year={2024},
      eprint={2405.05663},
      archivePrefix={arXiv},
      primaryClass={cs.CV},
      url={https://arxiv.org/abs/2405.05663}, 
}

@article{Xiao_2023,
   title={Unsupervised Point Cloud Representation Learning With Deep Neural Networks: A Survey},
   volume={45},
   ISSN={1939-3539},
   url={http://dx.doi.org/10.1109/TPAMI.2023.3262786},
   DOI={10.1109/tpami.2023.3262786},
   number={9},
   journal={IEEE Transactions on Pattern Analysis and Machine Intelligence},
   publisher={Institute of Electrical and Electronics Engineers (IEEE)},
   author={Xiao, Aoran and Huang, Jiaxing and Guan, Dayan and Zhang, Xiaoqin and Lu, Shijian and Shao, Ling},
   year={2023},
   month=sep, pages={11321–11339} }

@misc{yu2022pointbertpretraining3dpoint,
      title={Point-BERT: Pre-training 3D Point Cloud Transformers with Masked Point Modeling}, 
      author={Xumin Yu and Lulu Tang and Yongming Rao and Tiejun Huang and Jie Zhou and Jiwen Lu},
      year={2022},
      eprint={2111.14819},
      archivePrefix={arXiv},
      primaryClass={cs.CV},
      url={https://arxiv.org/abs/2111.14819}, 
}

@misc{pang2022maskedautoencoderspointcloud,
      title={Masked Autoencoders for Point Cloud Self-supervised Learning}, 
      author={Yatian Pang and Wenxiao Wang and Francis E. H. Tay and Wei Liu and Yonghong Tian and Li Yuan},
      year={2022},
      eprint={2203.06604},
      archivePrefix={arXiv},
      primaryClass={cs.CV},
      url={https://arxiv.org/abs/2203.06604}, 
}

@misc{zhang2022pointm2aemultiscalemaskedautoencoders,
      title={Point-M2AE: Multi-scale Masked Autoencoders for Hierarchical Point Cloud Pre-training}, 
      author={Renrui Zhang and Ziyu Guo and Rongyao Fang and Bin Zhao and Dong Wang and Yu Qiao and Hongsheng Li and Peng Gao},
      year={2022},
      eprint={2205.14401},
      archivePrefix={arXiv},
      primaryClass={cs.CV},
      url={https://arxiv.org/abs/2205.14401}, 
}

@misc{yang2021sanetshuffleattentiondeep,
      title={SA-Net: Shuffle Attention for Deep Convolutional Neural Networks}, 
      author={Qing-Long Zhang Yu-Bin Yang},
      year={2021},
      eprint={2102.00240},
      archivePrefix={arXiv},
      primaryClass={cs.CV},
      url={https://arxiv.org/abs/2102.00240}, 
}

@misc{xie2018learningdescriptornetworks3d,
      title={Learning Descriptor Networks for 3D Shape Synthesis and Analysis}, 
      author={Jianwen Xie and Zilong Zheng and Ruiqi Gao and Wenguan Wang and Song-Chun Zhu and Ying Nian Wu},
      year={2018},
      eprint={1804.00586},
      archivePrefix={arXiv},
      primaryClass={cs.CV},
      url={https://arxiv.org/abs/1804.00586}, 
}

@inproceedings{Rusu2011PCL,
  title     = {3D is here: Point Cloud Library (PCL)},
  author    = {Rusu, Radu Bogdan and Cousins, Steve},
  booktitle = {IEEE International Conference on Robotics and Automation (ICRA)},
  pages     = {1--4},
  year      = {2011},
  doi       = {10.1109/ICRA.2011.5980567}
}

@misc{PCL_SOR_ROR,
  title        = {Point Cloud Library (PCL) Filters: Statistical and Radius Outlier Removal},
  author       = {Rusu, Radu Bogdan and PCL Contributors},
  howpublished = {\url{https://pointclouds.org/documentation/}},
  note         = {Accessed: 2025-09-16}
}

@article{Fischler1981RANSAC,
  title   = {Random Sample Consensus: A Paradigm for Model Fitting with Applications to Image Analysis and Automated Cartography},
  author  = {Fischler, Martin A. and Bolles, Robert C.},
  journal = {Communications of the ACM},
  volume  = {24},
  number  = {6},
  pages   = {381--395},
  year    = {1981},
  doi     = {10.1145/358669.358692}
}

@inproceedings{Chetverikov2002TICP,
  title     = {The Trimmed Iterative Closest Point Algorithm},
  author    = {Chetverikov, Dmitry and Svirko, Dmitry and Stepanov, Dmitry and Krsek, Pavel},
  booktitle = {Proceedings of the 16th International Conference on Pattern Recognition (ICPR)},
  pages     = {545--548},
  volume    = {3},
  year      = {2002},
  doi       = {10.1109/ICPR.2002.1047997}
}

@inproceedings{Alexa2003PSS,
  title     = {Computing and Rendering Point Set Surfaces},
  author    = {Alexa, Marc and Behr, Johannes and Cohen-Or, Daniel and Fleishman, Shachar and Levin, David and Silva, Claudio T.},
  booktitle = {ACM SIGGRAPH},
  pages     = {141--150},
  year      = {2003},
  doi       = {10.1145/1201775.882293}
}

@article{Zeng2019GraphDenoise,
  title   = {Feature-Preserving Point Cloud Denoising via Graph Laplacian Regularization},
  author  = {Zeng, Yue and Cheung, Gene and Ng, Michael K. and Pang, John},
  journal = {IEEE Transactions on Image Processing},
  volume  = {29},
  pages   = {3474--3489},
  year    = {2019},
  doi     = {10.1109/TIP.2019.2962410}
}

@article{Bouaziz2013L0,
  title   = {$\ell_{0}$-based Point Set Denoising},
  author  = {Bouaziz, Sofien and Tagliasacchi, Andrea and Pauly, Mark},
  journal = {Computer Graphics Forum},
  volume  = {32},
  number  = {2pt2},
  pages   = {230--241},
  year    = {2013},
  doi     = {10.1111/cgf.12044}
}

@inproceedings{Rakotosaona2019PointCleanNet,
  title     = {PointCleanNet: Learning to Denoise and Remove Outliers from Dense Point Clouds},
  author    = {Rakotosaona, Marie-Julie and de Vita, Manuele and Litany, Or and Guerrero, Paul and Mitra, Niloy and Bronstein, Michael M.},
  booktitle = {Computer Graphics Forum (Proc. Eurographics)},
  volume    = {38},
  number    = {1},
  pages     = {234--245},
  year      = {2019},
  doi       = {10.1111/cgf.13525}
}

@article{Knapitsch2017,
    author    = {Arno Knapitsch and Jaesik Park and Qian-Yi Zhou and Vladlen Koltun},
    title     = {Tanks and Temples: Benchmarking Large-Scale Scene Reconstruction},
    journal   = {ACM Transactions on Graphics},
    volume    = {36},
    number    = {4},
    year      = {2017},
}

@misc{morshuis2024segmentationguidedmrireconstructionmeaningfully,
      title={Segmentation-guided MRI reconstruction for meaningfully diverse reconstructions}, 
      author={Jan Nikolas Morshuis and Matthias Hein and Christian F. Baumgartner},
      year={2024},
      eprint={2407.18026},
      archivePrefix={arXiv},
      primaryClass={eess.IV},
      url={https://arxiv.org/abs/2407.18026}, 
}

@article{Ma_2024,
   title={Segment anything in medical images},
   volume={15},
   ISSN={2041-1723},
   url={http://dx.doi.org/10.1038/s41467-024-44824-z},
   DOI={10.1038/s41467-024-44824-z},
   number={1},
   journal={Nature Communications},
   publisher={Springer Science and Business Media LLC},
   author={Ma, Jun and He, Yuting and Li, Feifei and Han, Lin and You, Chenyu and Wang, Bo},
   year={2024},
   month=jan }

@misc{li2024anatomaskenhancingmedicalimage,
      title={AnatoMask: Enhancing Medical Image Segmentation with Reconstruction-guided Self-masking}, 
      author={Yuheng Li and Tianyu Luan and Yizhou Wu and Shaoyan Pan and Yenho Chen and Xiaofeng Yang},
      year={2024},
      eprint={2407.06468},
      archivePrefix={arXiv},
      primaryClass={cs.CV},
      url={https://arxiv.org/abs/2407.06468}, 
}

@misc{zhou2025distillationlearningguidedimage,
      title={Distillation Learning Guided by Image Reconstruction for One-Shot Medical Image Segmentation}, 
      author={Feng Zhou and Yanjie Zhou and Longjie Wang and Yun Peng and David E. Carlson and Liyun Tu},
      year={2025},
      eprint={2408.03616},
      archivePrefix={arXiv},
      primaryClass={eess.IV},
      url={https://arxiv.org/abs/2408.03616}, 
}

@misc{cen2025segment3dgaussians,
      title={Segment Any 3D Gaussians}, 
      author={Jiazhong Cen and Jiemin Fang and Chen Yang and Lingxi Xie and Xiaopeng Zhang and Wei Shen and Qi Tian},
      year={2025},
      eprint={2312.00860},
      archivePrefix={arXiv},
      primaryClass={cs.CV},
      url={https://arxiv.org/abs/2312.00860}, 
}

@misc{dai2021can3dfast3dmedical,
      title={CAN3D: Fast 3D Medical Image Segmentation via Compact Context Aggregation}, 
      author={Wei Dai and Boyeong Woo and Siyu Liu and Matthew Marques and Craig B. Engstrom and Peter B. Greer and Stuart Crozier and Jason A. Dowling and Shekhar S. Chandra},
      year={2021},
      eprint={2109.05443},
      archivePrefix={arXiv},
      primaryClass={eess.IV},
      url={https://arxiv.org/abs/2109.05443}, 
}

@misc{xu2023stop3dtargetreconstruction,
      title={A One Stop 3D Target Reconstruction and multilevel Segmentation Method}, 
      author={Jiexiong Xu and Weikun Zhao and Zhiyan Tang and Xiangchao Gan},
      year={2023},
      eprint={2308.06974},
      archivePrefix={arXiv},
      primaryClass={cs.CV},
      url={https://arxiv.org/abs/2308.06974}, 
}

@misc{sun20233dfusionrealtime3dobject,
      title={3DFusion, A real-time 3D object reconstruction pipeline based on streamed instance segmented data}, 
      author={Xi Sun and Derek Jacoby and Yvonne Coady},
      year={2023},
      eprint={2311.06659},
      archivePrefix={arXiv},
      primaryClass={cs.CV},
      url={https://arxiv.org/abs/2311.06659}, 
}

@misc{pan2023multidimensionunifiedswintransformer,
      title={Multi-dimension unified Swin Transformer for 3D Lesion Segmentation in Multiple Anatomical Locations}, 
      author={Shaoyan Pan and Yiqiao Liu and Sarah Halek and Michal Tomaszewski and Shubing Wang and Richard Baumgartner and Jianda Yuan and Gregory Goldmacher and Antong Chen},
      year={2023},
      eprint={2309.01823},
      archivePrefix={arXiv},
      primaryClass={eess.IV},
      url={https://arxiv.org/abs/2309.01823}, 
}

@article{article,
author = {Li, Shuaiyong and Zhang, Xuyuntao and Zhang, Chao and Fu, Shenghao and Zhang, Sai},
year = {2024},
month = {06},
pages = {127945},
title = {Sparse Multi-view Image Clustering with Complete Similarity Information},
volume = {596},
journal = {Neurocomputing},
doi = {10.1016/j.neucom.2024.127945}
}

@misc{yan2024livelearncontinualaction,
      title={Live and Learn: Continual Action Clustering with Incremental Views}, 
      author={Xiaoqiang Yan and Yingtao Gan and Yiqiao Mao and Yangdong Ye and Hui Yu},
      year={2024},
      eprint={2404.07962},
      archivePrefix={arXiv},
      primaryClass={cs.CV},
      url={https://arxiv.org/abs/2404.07962}, 
}

@inproceedings{mauro2013overlapping,
  author = {Mauro, Massimo and Riemenschneider, Hayko and Van Gool, Luc and Leonardi, Riccardo},
  title = {Overlapping Camera Clustering Through Dominant Sets for Scalable 3D Reconstruction},
  booktitle = {Proceedings of the British Machine Vision Conference (BMVC)},
  year = {2013},
  month = {September},
  doi = {10.5244/C.27.120}
}

@inproceedings{khan2019fast,
  author = {Khan, Sahib and Bianchi, Tiziano},
  title = {Fast Image Clustering Based on Camera Fingerprint Ordering},
  booktitle = {Proceedings of the IEEE International Conference on Multimedia and Expo (ICME)},
  year = {2019},
  month = {July},
  pages = {766-771},
  doi = {10.1109/ICME.2019.00137}
}

@misc{thomas2019kpconvflexibledeformableconvolution,
      title={KPConv: Flexible and Deformable Convolution for Point Clouds}, 
      author={Hugues Thomas and Charles R. Qi and Jean-Emmanuel Deschaud and Beatriz Marcotegui and François Goulette and Leonidas J. Guibas},
      year={2019},
      eprint={1904.08889},
      archivePrefix={arXiv},
      primaryClass={cs.CV},
      url={https://arxiv.org/abs/1904.08889}, 
}

@misc{liu2019relationshapeconvolutionalneuralnetwork,
      title={Relation-Shape Convolutional Neural Network for Point Cloud Analysis}, 
      author={Yongcheng Liu and Bin Fan and Shiming Xiang and Chunhong Pan},
      year={2019},
      eprint={1904.07601},
      archivePrefix={arXiv},
      primaryClass={cs.CV},
      url={https://arxiv.org/abs/1904.07601}, 
}

@article{Guo_2021,
   title={PCT: Point cloud transformer},
   volume={7},
   ISSN={2096-0662},
   url={http://dx.doi.org/10.1007/s41095-021-0229-5},
   DOI={10.1007/s41095-021-0229-5},
   number={2},
   journal={Computational Visual Media},
   publisher={Tsinghua University Press},
   author={Guo, Meng-Hao and Cai, Jun-Xiong and Liu, Zheng-Ning and Mu, Tai-Jiang and Martin, Ralph R. and Hu, Shi-Min},
   year={2021},
   month=apr, pages={187–199} }

@misc{yu2018punetpointcloudupsampling,
      title={PU-Net: Point Cloud Upsampling Network}, 
      author={Lequan Yu and Xianzhi Li and Chi-Wing Fu and Daniel Cohen-Or and Pheng-Ann Heng},
      year={2018},
      eprint={1801.06761},
      archivePrefix={arXiv},
      primaryClass={cs.CV},
      url={https://arxiv.org/abs/1801.06761}, 
}

@misc{li2019puganpointcloudupsampling,
      title={PU-GAN: a Point Cloud Upsampling Adversarial Network}, 
      author={Ruihui Li and Xianzhi Li and Chi-Wing Fu and Daniel Cohen-Or and Pheng-Ann Heng},
      year={2019},
      eprint={1907.10844},
      archivePrefix={arXiv},
      primaryClass={cs.CV},
      url={https://arxiv.org/abs/1907.10844}, 
}

@article{rong2024repkpu,
  title={RepKPU: Point Cloud Upsampling with Kernel Point Representation and Deformation},
  author={Rong, Yifan and  },
  journal={Proceedings of the IEEE Conference on Computer Vision and Pattern Recognition (CVPR)},
  year={2024}
}

@misc{li2021pointcloudupsamplingdisentangled,
      title={Point Cloud Upsampling via Disentangled Refinement}, 
      author={Ruihui Li and Xianzhi Li and Pheng-Ann Heng and Chi-Wing Fu},
      year={2021},
      eprint={2106.04779},
      archivePrefix={arXiv},
      primaryClass={cs.CV},
      url={https://arxiv.org/abs/2106.04779}, 
}

@misc{oquab2023dinov2,
  title={DINOv2: Learning Robust Visual Features without Supervision},
  author={Oquab, Maxime and Darcet, Timothée and Moutakanni, Theo and Vo, Huy V. and Szafraniec, Marc and Khalidov, Vasil and Fernandez, Pierre and Haziza, Daniel and Massa, Francisco and El-Nouby, Alaaeldin and Howes, Russell and Huang, Po-Yao and Xu, Hu and Sharma, Vasu and Li, Shang-Wen and Galuba, Wojciech and Rabbat, Mike and Assran, Mido and Ballas, Nicolas and Synnaeve, Gabriel and Misra, Ishan and Jegou, Herve and Mairal, Julien and Labatut, Patrick and Joulin, Armand and Bojanowski, Piotr},
  journal={arXiv:2304.07193},
  year={2023}
}

@misc{simeoni2025dinov3,
  title={{DINOv3}},
  author={Sim{\'e}oni, Oriane and Vo, Huy V. and Seitzer, Maximilian and Baldassarre, Federico and Oquab, Maxime and Jose, Cijo and Khalidov, Vasil and Szafraniec, Marc and Yi, Seungeun and Ramamonjisoa, Micha{\"e}l and Massa, Francisco and Haziza, Daniel and Wehrstedt, Luca and Wang, Jianyuan and Darcet, Timoth{\'e}e and Moutakanni, Th{\'e}o and Sentana, Leonel and Roberts, Claire and Vedaldi, Andrea and Tolan, Jamie and Brandt, John and Couprie, Camille and Mairal, Julien and J{\'e}gou, Herv{\'e} and Labatut, Patrick and Bojanowski, Piotr},
  year={2025},
  eprint={2508.10104},
  archivePrefix={arXiv},
  primaryClass={cs.CV},
  url={https://arxiv.org/abs/2508.10104},
}

@misc{zhang2021viewguidedpointcloudcompletion,
      title={View-Guided Point Cloud Completion}, 
      author={Xuancheng Zhang and Yutong Feng and Siqi Li and Changqing Zou and Hai Wan and Xibin Zhao and Yandong Guo and Yue Gao},
      year={2021},
      eprint={2104.05666},
      archivePrefix={arXiv},
      primaryClass={cs.CV},
      url={https://arxiv.org/abs/2104.05666}, 
}

@misc{yu2021pointrdiversepointcloud,
      title={PoinTr: Diverse Point Cloud Completion with Geometry-Aware Transformers}, 
      author={Xumin Yu and Yongming Rao and Ziyi Wang and Zuyan Liu and Jiwen Lu and Jie Zhou},
      year={2021},
      eprint={2108.08839},
      archivePrefix={arXiv},
      primaryClass={cs.CV},
      url={https://arxiv.org/abs/2108.08839}, 
}

@misc{barron2021mipnerf,
      title={Mip-NeRF: A Multiscale Representation for Anti-Aliasing Neural Radiance Fields},
      author={Jonathan T. Barron and Ben Mildenhall and Matthew Tancik and Peter Hedman and Ricardo Martin-Brualla and Pratul P. Srinivasan},
      year={2021},
      eprint={2103.13415},
      archivePrefix={arXiv},
      primaryClass={cs.CV}
}

@InProceedings{Liu20neurips_sparse_nerf,
	Author = {Lingjie Liu and Jiatao Gu and Kyaw Zaw Lin and Tat-Seng Chua and Christian Theobalt},
	BookTitle = {Advances in Neural Information Processing Systems (NeurIPS)},
	Title = {Neural sparse voxel fields},
	Volume = {33},
	Year = {2020}
}

@article{Lindell20arxiv_AutoInt,
	Archiveprefix = {arXiv},
	Author = {David Lindell and Julien Martel and Gordon Wetzstein},
	Journal = {https://arxiv.org/abs/2012.01714},
	Title = {{AutoInt}: Automatic Integration for Fast Neural Volume Rendering},
	Year = {2020}
}

@article{barron2022mipnerf360,
    title={Mip-NeRF 360: Unbounded Anti-Aliased Neural Radiance Fields},
    author={Jonathan T. Barron and Ben Mildenhall and 
            Dor Verbin and Pratul P. Srinivasan and Peter Hedman},
    journal={CVPR},
    year={2022}
}

@article{yariv2023bakedsdf,
  title={BakedSDF: Meshing Neural SDFs for Real-Time View Synthesis},
  author={Yariv, Lior and Hedman, Peter and Reiser, Christian and Verbin, Dor and Srinivasan, Pratul P and Szeliski, Richard and Barron, Jonathan T and Mildenhall, Ben},
  journal={arXiv preprint arXiv:2302.14859},
  year={2023}
}

@article{cao2024lightning,
  title={{Lightning NeRF}: Efficient Hybrid Scene Representation for Autonomous Driving},
  author={Cao, Junyi and Li, Zhichao and Wang, Naiyan and Ma, Chao},
  journal={arXiv preprint arXiv:2403.05907},
  year={2024}
}

@misc{shang2024jointsegmentationimagereconstruction,
      title={Joint Segmentation and Image Reconstruction with Error Prediction in Photoacoustic Imaging using Deep Learning}, 
      author={Ruibo Shang and Geoffrey P. Luke and Matthew O'Donnell},
      year={2024},
      eprint={2407.02653},
      archivePrefix={arXiv},
      primaryClass={eess.IV},
      url={https://arxiv.org/abs/2407.02653}, 
}

@misc{oechsle2021unisurf,
      title={UNISURF: Unifying Neural Implicit Surfaces and Radiance Fields for Multi-View Reconstruction}, 
      author={Michael Oechsle and Songyou Peng and Andreas Geiger},
      year={2021},
      eprint={2104.10078},
      archivePrefix={arXiv},
      primaryClass={cs.CV}
}

@article{wang2021neus,
  title={NeuS: Learning Neural Implicit Surfaces by Volume Rendering for Multi-view Reconstruction},
  author={Peng Wang and Lingjie Liu and Yuan Liu and Christian Theobalt and Taku Komura and Wenping Wang},
  journal={NeurIPS},
  year={2021}
}

@article{yariv2021volume,
  title={Volume Rendering of Neural Implicit Surfaces},
  author={Yariv, Lior and Gu, Jiatao and Kasten, Yoni and Lipman, Yaron},
  journal={NeurIPS},
  year={2021}
}

@article{meng_2023_neat,
	title={NeAT: Learning Neural Implicit Surfaces with Arbitrary Topologies from Multi-view Images},
	author={Meng, Xiaoxu and Chen, Weikai and Yang, Bo},
	journal={Proceedings of the IEEE/CVF Conference on Computer Vision and Pattern Recognition},
	month={June},
	year={2023}
}

@misc{waleed2024cameracalibrationgeometricconstraints,
      title={Camera Calibration through Geometric Constraints from Rotation and Projection Matrices}, 
      author={Muhammad Waleed and Abdul Rauf and Murtaza Taj},
      year={2024},
      eprint={2402.08437},
      archivePrefix={arXiv},
      primaryClass={cs.CV},
      url={https://arxiv.org/abs/2402.08437}, 
}

@misc{gong2025dinoslamdinoinformedrgbdslam,
      title={DINO-SLAM: DINO-informed RGB-D SLAM for Neural Implicit and Explicit Representations}, 
      author={Ziren Gong and Xiaohan Li and Fabio Tosi and Youmin Zhang and Stefano Mattoccia and Jun Wu and Matteo Poggi},
      year={2025},
      eprint={2507.19474},
      archivePrefix={arXiv},
      primaryClass={cs.CV},
      url={https://arxiv.org/abs/2507.19474}, 
}

@misc{ravi2024sam2segmentimages,
      title={SAM 2: Segment Anything in Images and Videos}, 
      author={Nikhila Ravi and Valentin Gabeur and Yuan-Ting Hu and Ronghang Hu and Chaitanya Ryali and Tengyu Ma and Haitham Khedr and Roman Rädle and Chloe Rolland and Laura Gustafson and Eric Mintun and Junting Pan and Kalyan Vasudev Alwala and Nicolas Carion and Chao-Yuan Wu and Ross Girshick and Piotr Dollár and Christoph Feichtenhofer},
      year={2024},
      eprint={2408.00714},
      archivePrefix={arXiv},
      primaryClass={cs.CV},
      url={https://arxiv.org/abs/2408.00714}, 
}

@phdthesis{valverde2025convex,
  title     = {Convex-Guided Outlier Removal for 3D Point Clouds},
  author    = {Valverde Guillen, Alexander G.},
  year      = {2025},
  school    = {University of California, Santa Cruz},
  publisher = {ProQuest Dissertations Publishing},
  note      = {ProQuest ID: 32043589},
  url       = {https://escholarship.org/uc/item/55h4j9gj}
}
}

\end{document}